%% file: main.tex
\documentclass[
  english,        
  font=times,     
  onecolumn,      
]{tumarticle}

\usepackage{lipsum}
\usepackage{listings}
\usepackage{xcolor}
\usepackage[section]{placeins}
\usepackage{amsmath}
\usepackage{makecell}
\numberwithin{figure}{section}
\numberwithin{table}{section}

\definecolor{codegreen}{rgb}{0,0.6,0}
\definecolor{codegray}{rgb}{0.5,0.5,0.5}
\definecolor{codepurple}{rgb}{0.58,0,0.82}
\definecolor{backcolour}{rgb}{0.95,0.95,0.92}

\lstdefinestyle{mystyle}{
    backgroundcolor=\color{backcolour},   
    commentstyle=\color{codegreen},
    keywordstyle=\color{magenta},
    numberstyle=\tiny\color{codegray},
    stringstyle=\color{codepurple},
    basicstyle=\ttfamily\footnotesize,
    breakatwhitespace=false,         
    breaklines=true,                 
    captionpos=b,                    
    keepspaces=true,                 
    numbers=left,                    
    numbersep=5pt,                  
    showspaces=false,                
    showstringspaces=false,
    showtabs=false,                  
    tabsize=2
}

\title{Autonomous Driving Simulator based on Neurorobotics Platform}

\author[affil=1]{Wei Cao}
\author[affil=1, email=liguo.zhou@tum.de]{Liguo Zhou}
\author[affil=1]{Yuhong Huang}
\author[affil=1]{Alois Knoll}

\affil[mark=1]{\theChairName, \theUniversityName}

\date{}

\begin{document}

\maketitle

\input{sections/0_abstract.tex}
\input{sections/1_introduction.tex}
\input{sections/2_configuration.tex}
\input{sections/3_construction.tex}
\input{sections/4_result.tex}
\input{sections/5_conclusion.tex}

\bibliographystyle{unsrt} 
\bibliography{bib}


\appendix
\input{sections/6_appendix.tex}

\end{document}

%% file: sections/0_abstract.tex
\begin{abstract}
There are many artificial intelligence algorithms for autonomous driving in the present market, but directly installing these algorithms on vehicles is unrealistic and expensive. At the same time, many of these algorithms need an environment to train and optimize. Simulation is a valuable and meaningful solution with training and testing functions, and it can say that simulation is a critical link in the autonomous driving world. There are also many different applications or systems of simulation from companies or academies such as SVL and Carla. These simulators flaunt that they have the closest real-world simulation, but their environment objects, such as pedestrians and other vehicles around the agent-vehicle, are already fixed programmed. They can only move along the pre-setting trajectory, or random numbers determine their movements. What is the situation when all environmental objects are also installed by Artificial Intelligence, or their behaviors are like real people or natural reactions of other drivers? This problem is a blind spot for most of the simulation applications, or these applications cannot be easy to solve this problem. The Neurorobotics Platform from the TUM team of Prof. Alois Knoll has the idea about "Engines" and "Transceiver Functions" to solve the multi-agents problem. This report will start with a little research on the Neurorobotics Platform and analyze the potential and possibility of developing a new simulator to achieve the true real-world simulation goal. Then based on the NRP-Core Platform, this initial development aims to construct an initial demo experiment. The consist of this report starts with the basic knowledge of NRP-Core and its installation, then focus on the explanation of the necessary components for a simulation experiment, at last, about the details of constructions for the autonomous driving system, which is integrated object detection function and autonomous driving control function. At the end will discuss the existing disadvantages and improvements of this autonomous driving system.

\end{abstract}

\keywords{Simulation, Neurorobotics
Platform, NRP-Core, Engines, Transceiver Functions, Autonomous Driving, Object Detection, PID Trajectory Control}

%% file: sections/1_introduction.tex
\section{Introduction}

\subsection{Motivation}

At present, there are many different Artificial Intelligence (AI) algorithms used for autonomous driving. Some algorithms are used to perceive the environment, such as object detection and semantic/instance segmentation. Some algorithms are dedicated to making the best trajectory strategy and control decisions based on the road environment. Others contribute to many different applications, e.g. path planning and parking. Simulation is the best cost-performance way to develop these algorithms before they are truly deployed to actual vehicles or robots. So, the performance of a simulation platform is influencing the performance of the AI algorithms. In the present market or business world, there are already a lot of different “real-world” simulation applications such as CARLA~\cite{carla} for simulating the algorithm for autonomous driving, AirSim~\cite{airsim} from Microsoft for autonomous vehicle and quadrotor and PTV Vissim~\cite{vissim} from Germany PTV Group for flexible traffic simulation. 

Although these simulators are dedicated to the “real world” simulation, they have more or less “unreal” problems on some sides in the process of simulation. For example, besides the problem about the unreal 3-D models and environment, these simulators have an obvious feature, these AI algorithms are only deployed to target experimental subjects, vehicles, or robots, and the environment such as other vehicles, motorbikes, and pedestrian looks very close to the “real” environment but actually these environmental subjects are already in advance pre-programmed and have a fix motion trail. The core problem of most of them focuses on basic information transmission. They only transfer the essential or necessary traffic information to the agent subject in the simulation. This transmission is one-way direction. Considering this situation, can let other subjects in this simulation have their own different AI algorithms at the same time that they can react to the agent’s behavior? In the future world, there would be not only one vehicle owning one algorithm from one company, but they must also have much interaction with other agents. The interaction between different algorithms can take which influence back on these algorithms, and this problem is also a blind point for many simulators.

This large range of interaction between lots of agents is the main problem that these applications should pay attention to and these existing applications do not have an efficient way to solve this problem. A simulation platform that is truly like the real world, whose environment is not only a fixed pre-definition program, the objects in the environment can make a relative objective interaction with vehicles with the testing autonomous driving algorithms and they can influence each other, the goal and concept is an intractable problem for the construction of a simulation platform. There is a platform called The Neurorobotics Platform (NRP) from the TUM team of Prof. Alois Knoll that provides a potential idea to solve this interaction problem. This research project focuses on preliminary implementation and searches for the possibility of solving the previously mentioned interaction problem.

\subsection{Neurorobotics Platform (NRP)}

\begin{figure}[h]
    \centering
    \includegraphics[width=\textwidth]{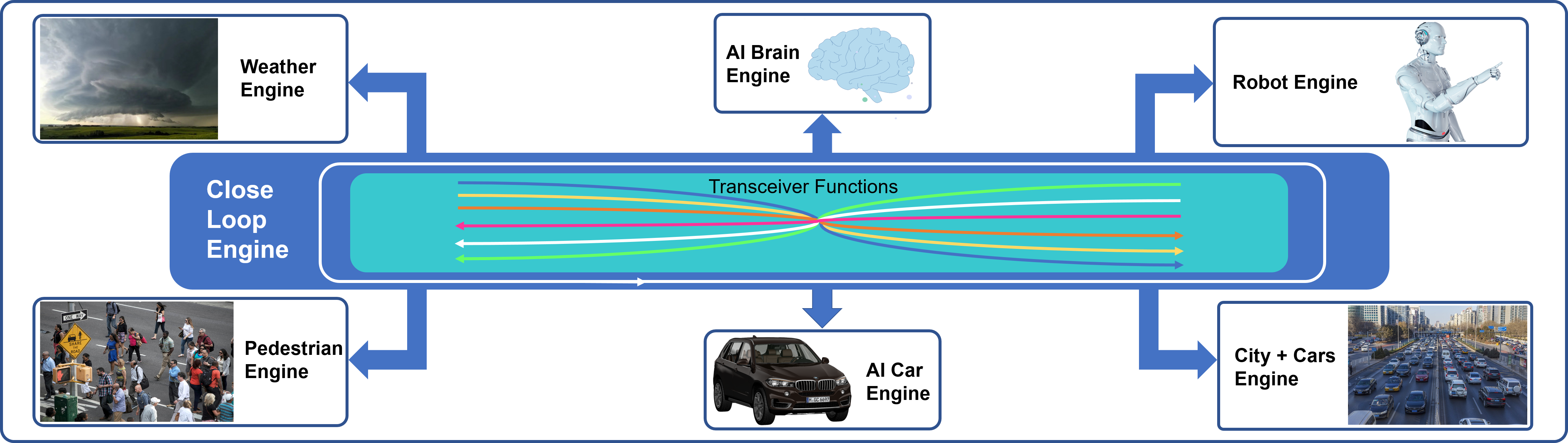}
    \caption{The base model of Neurorobotics Platform (NRP)}
    \label{fig1}
\end{figure}

Neurorobotics Platform~\cite{nrp} is an open-source integrative simulation framework platform developed by the group of the chair of Robotics, Artificial Intelligence and Real-Time Systems of the Technical University of Munich in the context of the Human Brain Project - a FET Flagship funded by the European Commission. The basic starting point of this platform enables to choose and test of different brain models (ranging from spiking neural networks to deep networks) for robots. This platform builds an efficient information transmission framework to let simulated agents interact with their virtual environment.

The new Version of NRP called NRP Core provides a new idea, which regards all the Participator in the Simulation-system as "Engines", just like the object in the programming language C++/python, the properties of the simulation participator such as the robot, autonomous-driving car, weather, or pedestrian and their "behaviors" would be completely constructed in their own "Engine"-object and let all the participates become a "real" object and can each other influence in the simulation world and they would not be a fix definite "Program". And the NRP-Platform is the most important transport median between these engines and they are called the Transceiver Function. It transmits the "Information" such as the image from the camera and sends the image to an autonomous-driving car and the same time would send other information to other engines by different transfer protocols such as JSON or ROS system. That means the transmission of information is highly real-time and lets the simulation world very close to the real world and it has high simulation potency, e.g. the platform sends the image information to the autonomous-driving car and lets the car computes the situation and makes the right strategy and rational decision, and at the same moment the environment-cars or "drivers" also get the location information from the autonomous-driving car and make their own decisions such like drive further or change velocity and lanes, and the same time these cars are influenced by the situation of the weather, e.g. in raining days the brake time of the car would be longer and let the decision making and object detection more significant.

NRP-core is mostly written in C++, with the Transceiver Function framework relying on Python for better usability. It guarantees a fully deterministic execution of the simulation, provided every simulator used is itself deterministic and works on the basis of controlled progression through time steps. Users should thus take note that event-based simulators may not be suitable for integration in NRP-core (to be analyzed on a case-by-case basis). Communications to and from NRP-core are indeed synchronous, and function calls are blocking; as such, the actual execution time of a simulation based on NRP-core will critically depend on the slowest simulator integrated therein. The aforementioned feature of the NRP-Core platform is significant to build multi-object which interact with other agencies in the simulation progress and lets the simulation be close to the real world. 

%% file: sections/2_configuration.tex
\section{NRP-Core configurations for simulation progress}

NRP-Core has many application scenarios for different demands of simulation situations. For a specific purpose, the model of NRP-Core can be widely different. This development for the Autonomous-driving benchmark focuses on the actual suggested development progress. It concentrates on the construction of the simulation application, the details of the operation mechanism of NRP-Core would not be discussed, and deep research in this development documentation, the principle of the operation mechanism can be found on the homepage of NRP-Core.

\subsection{Installation of NRP-Core and setting environment}

For the complete installation, refer to the homepage of the NRP-Core Platform by "Getting Started" under the page "Installation Instructions." This section lists only all the requirements for applying the autonomous driving simulator and benchmark.

\textbf{WARNING}: Previous versions of the NRP install forked versions of several libraries, notably NEST and Gazebo. Installing NRP-core in a system where a previous version of NRP is installed is known to cause conflicts. That will be strongly recommended not to install the last version at the same time.

\textbf{Operating System}: recommend on Ubuntu 20.04

Setting the Installation Environment:
To properly set the environment to run experiments with NRP-core, please make sure that it is added the lines below to your ~/.bashrc file.

\begin{lstlisting}[language=bash]
# Start setting environment
export NRP_INSTALL_DIR="/home/${USER}/.local/nrp" # The installation directory, which was given before
export NRP_DEPS_INSTALL_DIR="/home/${USER}/.local/nrp_deps"
export PYTHONPATH="${NRP_INSTALL_DIR}"/lib/python3.8/site-packages:"${NRP_DEPS_INSTALL_DIR}"/lib/python3.8/site-packages:$PYTHONPATH
export LD_LIBRARY_PATH="${NRP_INSTALL_DIR}"/lib:"${NRP_DEPS_INSTALL_DIR}"/lib:${NRP_INSTALL_DIR}/lib/nrp_gazebo_plugins:$LD_LIBRARY_PATH
export PATH=$PATH:"${NRP_INSTALL_DIR}"/bin:"${NRP_DEPS_INSTALL_DIR}"/bin
export GAZEBO_PLUGIN_PATH=${NRP_INSTALL_DIR}/lib/nrp_gazebo_plugins:${GAZEBO_PLUGIN_PATH}
. /usr/share/gazebo-11/setup.sh
. /opt/ros/noetic/setup.bash
. ${CATKIN_WS}/devel/setup.bash
# End of setting environment
\end{lstlisting}

\textbf{Dependency installation:}

\begin{lstlisting}[language=bash]
# Start of dependencies installation  
# Pistache REST Server  
sudo add-apt-repository ppa:pistache+team/unstable  

# Gazebo repository  
sudo sh -c 'echo "deb http://packages.osrfoundation.org/gazebo/ubuntu-stable `lsb_release -cs` main"> /etc/apt/sources.list.d/gazebo-stable.list'  
wget https://packages.osrfoundation.org/gazebo.key -O - | sudo apt-key add -  
      
sudo apt update  
sudo apt install git cmake libpistache-dev libboost-python-dev libboost-filesystem-dev libboost-numpy-dev libcurl4-openssl-dev nlohmann-json3-dev libzip-dev cython3 python3-numpy libgrpc++-dev protobuf-compiler-grpc libprotobuf-dev doxygen libgsl-dev libopencv-dev python3-opencv python3-pil python3-pip libgmock-dev  
   
# required by gazebo engine  
sudo apt install libgazebo11-dev gazebo11 gazebo11-plugin-base  
   
# Remove the flask if it was installed to ensure it is installed from pip  
sudo apt remove python3-flask python3-flask-cors  
# required by Python engine  
# If you are planning to use The Virtual Brain framework, you will most likely have to use flask version 1.1.4.  
# By installing flask version 1.1.4 markupsafe library (included with flask) has to be downgraded to version 2.0.1 to run properly with gunicorn  
# You can install that version with   
# pip install flask==1.1.4 gunicorn markupsafe==2.0.1  
pip install flask gunicorn  
   
# required by nest-server (which is built and installed along with nrp-core)  
sudo apt install python3-restrictedpython uwsgi-core uwsgi-plugin-python3   
pip install flask_cors mpi4py docopt  
   
# required by nrp-server, which uses gRPC python bindings  
pip install grpcio-tools pytest psutil docker  
   
# Required for using docker with ssh  
pip install paramiko  
     
# ROS, when not needed, can jump to the next step 
   
# Install ROS: follow the installation instructions: http://wiki.ros.org/noetic Installation/Ubuntu. To enable ros support in nrp on `ros-noetic-ros-base` is required.  
   
#Tell nrpcore where your catkin workspace is located: export a variable CATKIN_WS pointing to an existing catkin workspace root folder. If the variable does not exist, a new catkin workspace will be created at `${HOME}/catkin_ws`.  
   
# MQTT, if needed, see the homepage of NRP-Core

# End of dependencies installation 
\end{lstlisting}

\textbf{NRP installation:}

\begin{lstlisting}[language=bash]
# Start of installation  
git clone https://bitbucket.org/hbpneurorobotics/nrp-core.git  
cd nrp-core  
mkdir build  
cd build  
# See the section "Common NRP-core CMake options" in the documentation for the additional ways to configure the project with CMake  
cmake .. -DCMAKE_INSTALL_PREFIX="${NRP_INSTALL_DIR}" -DNRP_DEP_CMAKE_INSTALL_PREFIX="${NRP_DEPS_INSTALL_DIR}"  
mkdir -p "${NRP_INSTALL_DIR}"  
# the installation process might take some time, as it downloads and compiles Nest as well.
# If you haven't installed MQTT libraries, add ENABLE_MQTT=OFF definition to cmake (-DENABLE_MQTT=OFF).  
make  
make install  
# Just in case of wanting to build the documentation. Documentation can then be found in a new doxygen folder  
make nrp_doxygen  
# End of installation  
\end{lstlisting}

\textbf{Common NRP-core CMake options}:
Here is the list of the CMake options that can help modify the project configuration (turn on and off the support of some components and features).

\begin{itemize}
\item Developers options:
    \begin{itemize}
        \item COVERAGE enables the generation of the code coverage reports during the testing
        \item BUILD\_RST enables the generation of the reStructuredText source files from the Doxygen documentation
    \end{itemize}
    
\item Communication protocols options:
    \begin{itemize}
        \item ENABLE\_ROS enables compilation with ROS support;
        \item ENABLE\_MQTT enables compilation with the MQTT support.
    \end{itemize}

\item ENABLE\_SIMULATOR and BUILD\_SIMULATOR\_ENGINE\_SERVER options:
    \begin{itemize}
        \item ENABLE\_NEST and BUILD\_NEST\_ENGINE\_SERVER;
        \item ENABLE\_GAZEBO and BUILD\_GAZEBO\_ENGINE\_SERVER.
    \end{itemize}
\end{itemize}

The ENABLE\_SIMULATOR and BUILD\_SIMULATOR\_ENGINE\_SERVER flags allow disabling the compilation of those parts of nrp-core that depend on or install a specific simulator (eg. gazebo, nest).

The expected behavior for each of these pairs of flags is as follows:

\begin{itemize}
\item the NRPCoreSim is always built regardless of any of the flags values.

\item if ENABLE\_SIMULATOR is set to OFF:
    \begin{itemize}
    \item the related simulator won't be assumed to be installed in the system, ie. make won't fail if it isn't. Also it won't be installed in the compilation process if this possibility is available (as in the case of nest)
    \item The engines connected with this simulator won't be built (nor client nor server components)
    \item tests that would fail if the related simulator is not available won't be built
    \end{itemize}
\item if the ENABLE\_SIMULATOR is set to ON and BUILD\_SIMULATOR\_ENGINE\_SERVER is set to OFF: Same as above, but:
    \begin{itemize}
    \item the engine clients connected to this simulator will be built. This means that they should not depend on or link to any specific simulator
    \item the engine server-side components might or might not be built, depending on if the related simulator is required at compilation time
    \end{itemize}
\item if both flags are set to ON the simulator is assumed to be installed or it will be installed from the source if this option is available. All targets connected with this simulator will be built.
\end{itemize}

This flag system allows configuring the resulting NRP-Core depending on which simulators are available on the system, both for avoiding potential dependency conflicts between simulators and enforcing modularity, opening the possibility of having specific engine servers running on a different machine or inside containers.

\subsection{Introduction of basic components of simulation by NRP}

Some important elements for constructing a simulation example by the NRP platform are: Engines, Transceiver Function (TF) + Preprocessing Function (PF), Simulation Configuration JSON file, Simulation model file and DataPack, which are basic components of simulation progress. In this section, list and declare their definition, content and implementation.

\subsubsection{Engine}

Engines are a core aspect of the NRP-core framework. They run the actual simulation software (which can be comprised of any number of heterogeneous modules), with the Simulation Loop and TransceiverFunctions merely being a way to synchronize and exchange data between them. The data exchange is carried out through an engine client (see paragraph below). An Engine can run any type of software, from physics engines to brain simulators. The only requirement is that they should be able to manage progressing through time with fixed-duration time steps.

There are different engines already implemented in NRP-Core:
\begin{itemize}
\item Nest: two different implementations that integrate the NEST Simulator into NRP-core.
\item Gazebo: engine implementation for the Gazebo physics simulator.
\item PySim: engine implementation based on the Python JSON Engine wrapping different simulators (Mujoco, Opensim, and OpenAI) with a python API.
\item The Virtual Brain: engine implementation based on the Python JSON Engine and TVB Python API.
\end{itemize}
and so on are provided by NRP and as the first user-interested engines for research Spiking neural Networks and the like. These applications are distributed to the specific simulator. This platform provides also \textbf{Python JSON Engine}, this versatile engine enables users to execute a user-defined python script as an engine server, thus ensuring synchronization and enabling DataPack data transfer with the Simulation Loop process. It can be used to integrate any simulator with a Python API in an NRP-core experiment. This feature allows users to modular develop experiment agents in constructed simulation world and is flexible to manage plural objects with different behaviors and characters.

\subsubsection{DataPack and Construction format}

The carrier of Information which is transported between engines and lets engines with each other communicate is DataPack. By NRP are there three types of supported DataPack, all of them are simple objects which wrap around arbitrary data structures, one is \textbf{JSON DataPack}, second is \textbf{Protobuf DataPack} and another is \textbf{ROS msg DataPack}. They provide the necessary abstract interface, which is understood by all components of NRP-Core, while still allowing the passing of data in various formats. DataPack is also an important feature or property of a specific Engine, meaning the parameters and form of data of a specific DataPack be declared in the Engine (Example see section 3.4.2).

A DataPack consists of two parts:

\begin{itemize}
\item DataPack ID: which allows unique identify the object.
\item DataPack data: this is the data stored by the DataPack, which can be in the principle of any type.
\end{itemize}

DataPacks are mainly used by Transceiver functions to relay data between engines. Each engine type is designed to accept only datapacks of a certain type and structure. 

Every DataPack contains a DataPackIdentifier, which uniquely identifies the datapack object and allows for the routing of the data between transceiver functions, engine clients and engine servers. A datapack identifier consists of three fields:
\begin{itemize}
\item name - the name of the DataPack. It must be unique.
\item type - string representation of the DataPack data type. This field will most probably be of no concern for the users. It is set and used internally and is not in human-readable form.
\item engine name - the name of the engine to which the DataPack is bound.
\end{itemize}
DataPack is a template class with a single template parameter, which specifies the type of data contained by the DataPack. This DataPack data can be in the principle of any type. In practice, there are some limitations though, since DataPacks, which are C++ objects, must be accessible from TransceiverFunctions, which are written in Python. Therefore the only DataPack data types which can be actually used in NRP-core are those for which Python bindings are provided. It is possible for a DataPack to contain no data. This is useful, for example, when an Engine is asked for a certain DataPack but it is not able to provide it. In this case, an Engine can return an empty DataPack. This type of Datapack contains only a Datapack identifier and no data. Attempting to retrieve the data from an empty DataPack will result in an exception. A method "isEmpty" is provided to check whether a DataPack is empty or not before attempting to access its data:

\begin{lstlisting}[language=python]
if(not datapack.isEmpty()):  
    # It's safe to get the data  
    print(datapack.data)  
else:  
    # This will raise an exception  
    print(datapack.data)  
\end{lstlisting}

\begin{itemize}
\item The Format of getting DataPack from a particular Engine:

\begin{lstlisting}[language=python]
# Declare datapack with "datapack_name" name from engine "engine_name" as input using the @EngineDataPack decorator  
# The transceiver function must accept an argument with the same name as "keyword" in the datapack decorator  
  
@EngineDataPack(keyword="datapack", id=DataPackIdentifier("datapack_name", "engine_name"))  
@TransceiverFunction("engine_name")  
def transceiver_function(datapack):  
    print(datapack.data)  
   
# Multiple input datapacks from different engines can be declared  
@EngineDataPack(keyword="datapack1", id=DataPackIdentifier("datapack_name1", "engine_name1"))
@EngineDataPack(keyword="datapack2", id=DataPackIdentifier("datapack_name2", "engine_name2"))
@TransceiverFunction("engine_name1")  
def transceiver_function(datapack1, datapack2):  
    print(datapack1.data)  
    print(datapack2.data)  
\end{lstlisting}

PS: The details of two Decorators of TransceiverFunction see below in section 2.2.3.

\item The Format of setting information in DataPack and sending to particular Engine:
\begin{lstlisting}[language=python]
# NRP-Core expects transceiver functions to always return a list of datapacks   
@TransceiverFunction("engine_name")  
def transceiver_function():  
    datapack = JsonDataPack("datapack_name", "engine_name")  
    return [ datapack ]  
   
# Multiple datapacks can be returned  
  
@TransceiverFunction("engine_name")  
def transceiver_function():  
    datapack1 = JsonDataPack("datapack_name1", "engine_name")  
    datapack2 = JsonDataPack("datapack_name2", "engine_name")  
   
    return [ datapack1, datapack2 ] 
\end{lstlisting}

\end{itemize}

\subsubsection{Transceiver Function and Preprocessing Function}

\textbf{1. Transceiver Function}

Transceiver Functions are user-defined Python functions that take the role of transmitting DataPacks between engines. They are used in the architecture to convert, transform or combine data from one or multiple engines and relay it to another. 

The definition of a Transceiver Function must use Decorator before the user-defined “def” transceiver function, which means: Sending the DataPack to the target Engine: 
\begin{lstlisting}[language=python]
@TransceiverFunction("engine_name")
\end{lstlisting}
To request datapacks from engines, additional decorators can be prepended to the Transceiver Function, with the form (Attention: Receive-Decorator must be in the front of TransceiverFunction):
\begin{lstlisting}[language=python]
@EngineDataPack(keyword_datapack, id_datapack)
\end{lstlisting}

\begin{itemize} 
\item keyword\_datapack: user-defined new data name of DataPacks, this keyword is used as Input to Transceiver Function.

\item id\_datapack: 	the id of from particular Engine received DataPack, 
“DataPack ID” = “DataPack Name” + “Engine Name”
(Examples see 2.2.2)
\end{itemize}
\textbf{2. Preprocessing Function}

Preprocessing Function is very similar to Transceiver Function but has different usage. Preprocessing Functions are introduced to optimize expensive computations on DataPacks attached to a single engine. In some cases, there might be necessary to apply the same operations on a particular DataPack in multiple Transceiver Functions. An example of this might be applying a filter to a DataPack containing an image from a physics simulator. In order to allow to execute this operation just once and let other TFs access the processed DataPack data, PreprocessingFunctions (PFs) are introduced.

They show two main differences with respect to Transceiver Functions:
\begin{itemize} 
\item Their output datapacks are not sent to the corresponding Engines, they are kept in a local datapack cache and can be used as input in TransceiverFunctions
\item PFs just can take input DataPacks from the Engine they are linked to
\end{itemize}

The format of Preprocessing Function is similar to Transceiver Function:
\begin{lstlisting}[language=python]
@PreprocessingFunction("engine_name")  
@PreprocessedDataPack(keyword_datapack, id_datapack) 
\end{lstlisting}

These Decorators “@PreprocessingFunction” and “@PreprocessedDataPack” must be used in Preprocessing Functions. Since the output of Preprocessing Function is stored in the local cache and does not need to process on the Engine Server side, Preprocessing Function can return any type of DataPack without restrictions.

\subsubsection{Simulation Configuration Json file}

The details of configuration information for any simulation with Engines and Transceiver Functions are stored in a single JSON file, this file contains the objects of engines, Transceiver functions, and also their important necessary parameters to initialize and execute a simulation. This file is usually written in the “example\_simlation.json” file. 

The JSON format is here a JSON schema, which is highly readable and offers similar capabilities as XML Schema. The advantage of composability and inheritance allows the simulation to use reference keywords to definite the agent and to validate inheritance by referring to other schemas. That means that the same basement of an engine can at the same time create plural agents or objects with only different identify IDs.

\textbf{1. Simulation Parameters}

For details, see appendix Table~\ref{a1}: Simulation configuration parameter.

\textbf{2. Example form}

\begin{lstlisting}[language=XML]
{  
    "SimulationName": "example_simulation",  
    "SimulationDescription": "Launch two python engines. "
    "SimulationTimeout": 1,  
    "EngineConfigs":   
    [  
        {  
            "EngineType": "python_json",  
            "EngineName": "python_1",  
            "PythonFileName": "engine_1.py"  
        },  
        {  
            "EngineType": "python_json",  
            "EngineName": "python_2",  
            "PythonFileName": "engine_2.py"  
        }  
    ],  
    "DataPackProcessingFunctions":  
    [  
        {  
            "Name": "tf_1",  
            "FileName": "tf_1.py"  
        }  
    ]  
}   
\end{lstlisting}

\begin{itemize}
\item EngineConfigs: this section list all the engines are participating in the simulation progress. There are some important parameters should be declared:
    \begin{itemize}
    \item EngineType: which type of engine is used for this validated engine, e.g., gazebo engine, python JSON engine
    \item EngineName: user-defined unit identification name for the validated engine
    \item Other Parameters: These Parameters should be declared according to the type of engines (details see appendix Table~\ref{a2}: Engine Base Parameter)
        \begin{itemize}
        \item Python Json engine: “PythonFileName” – reference base python script for validated engine
        \item Gazebo engine: see in section
        \end{itemize}
    \end{itemize}

\item DataPackProcessingFunctions: this section lists all the Transceiver functions validated in simulation progress. Mostly are there two parameters that should be declared:
    \begin{itemize}
    \item Name: user-defined identification name for validated Transceiver Function
    \item FileName: which file as reference base python script to validate Transceiver Function
    \end{itemize}

\item Other Simulation Parameters: see section 2.2.4 – 1. Simulation Parameters

\item Launch a simulation: This simulation configuration JSON file is also the launch file and uses the NRP command to start a simulation experiment with the following command:

\begin{lstlisting}[language=bash]
NRPCoreSim -c user_defined_simulation_config.json
\end{lstlisting}

\end{itemize} 
Tip: In a user-defined simulation, the folder can simultaneously exist many different named configuration JSON files. It is very useful to config the target engine or Transceiver Functions that which user wants to launch and test with. To start and launch the target simulation experiment, just choose the corresponding configuration file.

\subsubsection{Simulation model file}

In this experiment for Autonomous driving on the NRP platform Gazebo physics simulator~\cite{gazebo} is the world description simulator. For the construction of the simulation, the world can use the “SDF” file based on XML format to describe all the necessary information about 3D models in a file, e.g. sunlight, environment, friction, wind, landform, robots, vehicles, and other physics objects. This file can in detail describe the static or dynamic information of the robot, the relative position and motion information, the declaration of sensor or control plugins, and so on. And Gazebo is a simulator that has a close correlation to the ROS system and provides simulation components for ROS, so the ROS system describes many similar details about the construction of SDF file~\cite{urdf}.

According to XML format label to describe components of the simulation world and construct the dependence relationship of these components:

\begin{itemize}
\item World Label
\begin{lstlisting}[language=xml]
<sdf version='1.7'>  
    <world name='default'>  
        ........  
    </world>  
</sdf>
\end{lstlisting}
All the components and their labels should be under <world> label.

\item Model Labels
\begin{lstlisting}[language=xml]
<model name='model_name'>  
    <pose>0 0 0 0 -0 0</pose>  
    <link name='road map'>  
        .........  
    </link>  
    <plugin name='link_plugin' filename='NRPGazeboGrpcLinkControllerPlugin.so'/> 
    </plugin>  
</model>
\end{lstlisting}
The Description is under label <model>, and importantly if user will use a plugin such as the control-plugin or sensor-plugin (camera or lidar), this <plugin> label must be set under the corresponding <model> label. Under <link> label describes the model physics features like <collision>, <visual>, <joint>, and so on.

\item 3-D models – mesh files

Gazebo requires that mesh files be formatted as STL, Collada, or OBJ, with Collada and OBJ being the preferred formats. Blow lists the file suffixes to the corresponding mesh file format.

\centerline{Collada - .dae \quad OBJ - .obj \quad STL - .stl}

Tip: Collada and OBJ file formats allow users to attach materials to the meshes. Use this mechanism to improve the visual appearance of meshes.
        
Mesh file should be declared under a needed label like <visual> or <collision> with layer structure with <geometry> - <mesh> - <uri> (Uri can be absolute or relative file path):
\begin{lstlisting}[language=xml]
<geometry>  
    <mesh>  
        <uri>xxxx/xxxx.dae</uri>  
    </mesh>  
</geometry>
\end{lstlisting}
\end{itemize}

%% file: sections/3_construction.tex
\section{Simulation Construction on NRP-Core}

Based on the steps for configuring a simulation on the NRP-Core platform, the autonomous driving benchmark can now be implemented with the components mentioned above, from 3D models to communicating mechanisms. This section will introduce the requirements of the autonomous driving application, and second will analyze the corresponding components and their functions. The third is the concrete implementation of these requirements. 

Second, this project will also research the possibility of achieving modular development for multi-agents on the NRP platform, comparing it with other existing and widely used systems, and analyzing the simulation performance according to the progress result.

\subsection{Analysis of requirements for autonomous driving application}

An application to achieve the goal of testing the performance of autonomous driving algorithms can refer to different aspects. The reason is that autonomous driving can integrate different algorithms such as computer vision, object detection, decision-making and trajectory planning, vehicle control, or Simultaneous localization and mapping. The concept and final goal of the application are to build a real-world simulation that integrates multi-agents, different algorithms, and corresponding evaluation systems to the performance of the autonomous driving vehicle. But that first needs many available, mature, and feasible algorithms. Second, the construction of world 3D models is a big project. And last, the evaluation system is based on the successful operation of the simulation. So the initial construction of the application will focus on the base model of the communication mechanism to first achieve the communication between the single agent and object-detection algorithm under the progress of NRP-Core. And for vehicle control algorithm reacts logically based on the object detection and generates feasible control commands, in this project will skip this step and give a specific trajectory, that let the vehicle along this trajectory move. 

Requirements of implementation:
\begin{itemize}
    \item Construction of the base model frame for communication between the Gazebo simulator, object-detection algorithm, and control unit.
    \item Selection of feasible object-detection algorithm
    \item Simple control system for autonomous movement of high accuracy physical vehicle model
\end{itemize}

\subsection{Object detection algorithm and YOLO v5 Detector Python Class}

According to the above analysis, the requirements of the application should choose an appropriate existing object detection algorithm as the example to verify the communication mechanism of the NRP platform and at the same time to optimize performance. 

On the research of existing object detection algorithms from base Alex-Net for image classification~\cite{alexnet} and CNN-Convolution neural network for image recognition~\cite{vgg}, the optimized neural network ResNet~\cite{resnet} and SSD neural network for multi-box Detector~\cite{ssd} and in the end the YOLOv5 neural network~\cite{yolov5}, YOLOv5 has high performance on the object detection and its advantage by efficient handling of frame image on real-time let this algorithm also be meaningful as a reference to test other object-detection algorithms. Considering the requirements of autonomous driving is YOLOv5 also a suitable choice as the experimental object-detection algorithm to integrate into the NRP platform.

Table Notes:
\begin{itemize} 
\item All checkpoints are trained to 300 epochs with default settings and hyperparameters.
\item mAPval values are for single-model single-scale on COCO val2017 dataset. Reproduced by python val.py --data coco.yaml --img 640 --conf 0.001 --iou 0.65
\item Speed averaged over COCO val images using a AWS p3.2xlarge instance. NMS times (~1 ms/img) not included.Reproduce by python val.py --data coco.yaml --img 640 --conf 0.25 --iou 0.45
\item TTA Test Time Augmentation includes reflection and scale augmentations.Reproduce by python val.py --data coco.yaml --img 1536 --iou 0.7 --augment
\end{itemize}

Requirements and Environment for YOLOv5:
\begin{itemize}
\item Quick link for YOLOv5 documentation : YOLOv5 Docs~\cite{yolov5doc}
\item Environment requirements: Python >= 3.7.0 version and PyTorch~\cite{pytorch} >= 1.7
\item Integration of original initial trained YOLOv5 neural network parameters, the main backbone has no changes compared to the initial version
\end{itemize}

Based on the original execute-python file “detect.py” has another python file “Yolov5Detector.py” with a self-defined Yolov5Detector class interface written in the “YOLOv5” package. To use YOLO v5 should in main progress validate the YOLO v5 class, second use warm-up function “\textbf{detectorWarmUp()}” to initiate the neural network. And “\textbf{detectImage()}” is the function that sends the image frame to the main predict detection function and will finally return the detected image with bounding boxes in NumPy format.

\subsection{3D-Models for Gazebo simulation world}

According to the performance of the Gazebo is the scope of the base environment world not suitable to use a large map. On the basic test of different sizes of the map of Garching-area is the environment world model recommends encircling the area of Parkring in Garching-Hochbrück. This map model is based on the high-accuracy satellite generated and is very similar to the origin location. And by the simulation progress, the experimental vehicle moves around the main road of Parkring. 

\begin{figure}[ht]
    \centering
    \subfloat[][Parkring Garching Hochbrueck high accuracy map model]{
    \includegraphics[height=5cm]{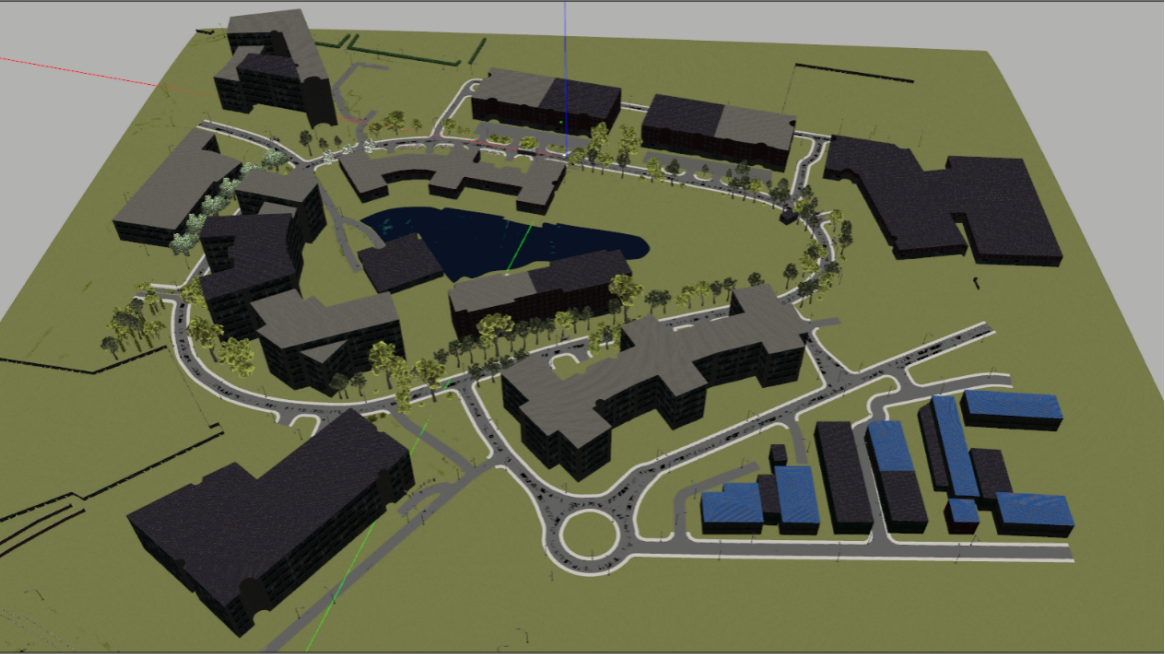}
    \label{parkring}
    }
    \subfloat[][Experiment vehicle for simulation]{
    \includegraphics[height=5cm]{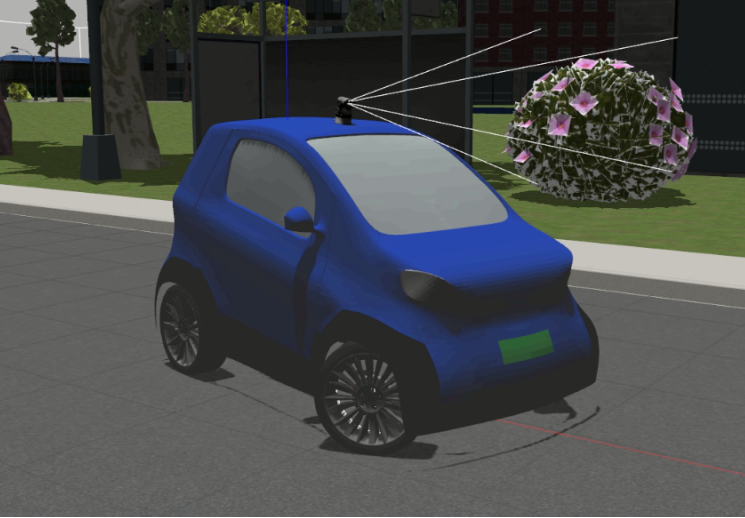}
    \label{vehicle}
    }
    \caption{}
\end{figure}

The experimental vehicle is also a high detail modeling vehicle model with independently controllable steerings for diversion control of two front wheels, free front, and rear wheels, and a high-definition camera. For the rebuilding of these models, the belonging relationship for each mode should be declared in the SDF file. In the SDF file are these models including base-chassis, steerings, wheels, and camera as “Link” of the car “model” under the <model> label with a user-defined unique name. Attention, the name of models or links must be specific and has no same name as other objects. The below shows the base architecture frame to describe the physical relationship of the whole vehicle in the SDF file:

\begin{lstlisting}[language=xml]
<model name='smart_car'>  
    <link name='base_link'>  
       .......  
    </link>  
    <link name='eye_vision_camera'>  
       .......  
    </link>  
    <joint name='eye_vision_camera_joint' type='revolute'>  
       <parent>base_link</parent>  
       <child>eye_vision_camera</child>  
       ......  
    </joint>  
    <link name='front_left_steering_link'>  
        .......  
    </link>  
    <joint name='front_left_steering_joint' type='revolute'>  
       <parent>base_link</parent>  
       <child>front_left_steering_link</child>  
        .......  
    </joint>  
    ......  
</model> 
\end{lstlisting}

\textbf{1. Description of Labels~\cite{urdf}:}

\begin{itemize} 
\item <link> 		---	 	The corresponding model as a component from the entirety model
\item <joint>		---		Description of relationship between link-components
\item <joint type> 	--- 		Type of the joint:
    \begin{itemize}
    \item revolute — a hinge joint that rotates along the axis and has a limited range specified by the upper and lower limits.
    \item continuous — a continuous hinge joint that rotates around the axis and has no upper and lower limits.
    \item prismatic — a sliding joint that slides along the axis, and has a limited range specified by the upper and lower limits.
    \item fixed — this is not a joint because it cannot move. All degrees of freedom are locked. This type of joint does not require the <axis>, <calibration>, <dynamics>, <limits> or <safety\_controller>.
    \item floating — this joint allows a motion for all 6 degrees of freedom.
    \item planar — this joint allows motion in a plane perpendicular to the axis.
    \end{itemize}
\item <parent>/<child>  ---	the secondary label as element of <joint> label
                        
                        ---	declaration for the belonging relationship of referring “links”
\end{itemize}

The mesh file “vehicle\_body.dae” (shown in Fig.~\ref{vehicle} the blue car body) is used for the base-chassis of the experiment vehicle under <link name=‘base\_link’> label. And the mesh file “wheel.dae” is used for the rotatable vehicle wheels under <link name=’ front\_left\_wheel\_link’> and the other three similar link labels. And for steering models, <cylinder> labels are used to simply generate length – 0.01m + height radius 0.1m cylinder as the joint elements between wheels and chassis.

\textbf{2. Sensor Label:}

In the Gazebo simulator to activate the camera function, the camera model should under the “camera link” label declare a new secondary “sensor label” - <sensor> with “name” and “type=camera” elements. And the detailed construction for the camera sensor seeing blow scripts:

\begin{lstlisting}[language=xml]
<sensor name='camera' type='camera'>  
   <pose>0 0 0.132 0 -0.174 0</pose>  
   <topic>/smart/camera</topic>  
   <camera>  
      <horizontal_fov>1.57</horizontal_fov>  
      <image>  
         <width>736</width>  
         <height>480</height>  
      </image>  
      <clip>  
         <near>0.1</near>  
         <far>100</far>  
      </clip>  
      <noise>  
         <type>gaussian</type>  
         <mean>0</mean>  
         <stddev>0.007</stddev>  
      </noise>  
   </camera>  
   <always_on>1</always_on>  
   <update_rate>30</update_rate>  
   <visualize>1</visualize>  
</sensor>
\end{lstlisting}

\begin{itemize}
\item <image>	---	this label defines the camera resolution ratio and this is regarded as the size of the frame-image that sends to the Yolo detector engine. According to the requirement of the YOLO detection algorithm, the width and height of the camera should be set as integral multiples by 32.
\end{itemize}

\subsection{Construction of Engines and Transceiver Functions}

\begin{figure}[h]
    \centering
    \includegraphics[width=\textwidth]{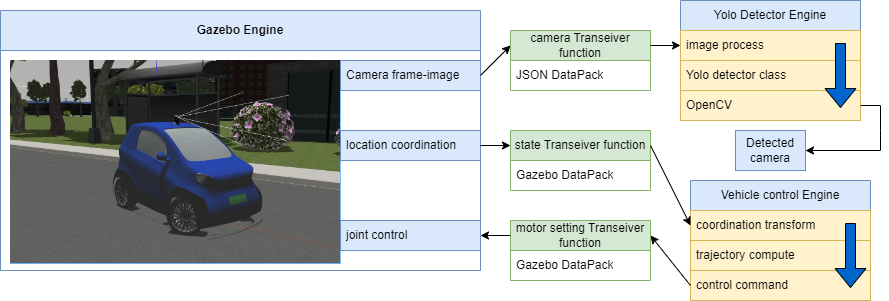}
    \caption{the system of autonomous driving on NRP}
    \label{system}
\end{figure}

The construction of the whole project regards as an experiment on the NRP platform, and as an experiment, the whole package of the autonomous driving benchmark is under the “nrp-core” path in the examples folder. According to bevor announced NRP components for a simulation experiment is the application also modular developed referring to requirements of autonomous driving benchmark application. And the whole system frame is shown in Fig.~\ref{system}. The construction of simulation would according to primary embrace two branches extend: 

\begin{itemize}
\item A close loop from the Gazebo engine to get the location information of the vehicle and sent to the Vehicle control engine depending on Gazebo DataPacks (Protobuf DataPack), then send the joint control command back to the Gazebo engine. 

\item An open loop from Gazebo engine to get camera information and sent to Yolo Detector Engine, final using OpenCV to show the detected frame-image as monitor window.
\end{itemize}

\subsubsection{Gazebo plugins}

Before the steps to acquire the different information must the corresponding plugins in SDF be declared. These plugins label are such as recognition-label to let Gazebo know what information and parameters should be sent or received and assigned. A set of plugins is provided to integrate the Gazebo in NRP-Core simulation. NRPGazeboCommunicationPlugin registers the engine with the SimulationManager and handles control requests for advancing the gazebo simulation or shutting it down. Its use is mandatory in order to run the engine. And there are two implementations of the Gazebo engine are provided. One is based on JSON over REST and another on Protobuf over gRPC. The latter performs much better and it is recommended. The gRPC implementation uses protobuf objects to encapsulate data exchanged between the Engine and TFs, whereas the JSON implementation uses nlohmann::json objects. Besides this fact, both engines are very similar in their configuration and behavior. The rest of the documentation below is implicitly referred to the gRPC implementation even though in most cases the JSON implementation shows no differences. The corresponding plugins are also based on Protobuf over the gRPC protocol. There are four plugins that would be applied in the SDF model world file:

\begin{itemize}

\item World communication plugin – NRPGazeboGrpcCommunicationPlugin

This plugin is the main communication plugin to set up a gRPC server and waits for NRP commands. It must be declared under the <world> label in the SDF file.
\begin{lstlisting}[language=xml]
<world name='default'>  
...  
    <plugin name="nrp_world_plugin" filename="NRPGazeboGrpcWorldPlugin.so"/>  
...  
</world> 
\end{lstlisting}

\item Activation of Camera sensor plugin – NRPGazeboGrpcCameraPlugin

This plugin is used to add a GazeboCameraDataPack datapack. In the SDF file, the plugin would be named “smart\_camera” (user-defined). This name can be accessed by TransceiverFunctions and get the corresponding information. This plugin must be declared under <sensor> label in the application under the camera sensor label:
\begin{lstlisting}[language=xml]
<sensor name='camera' type='camera'>  
   ...  
   <plugin name='smart_camera' filename='NRPGazeboGrpcCameraControllerPlugin.so'/>  
   ...  
</sensor> 
\end{lstlisting}

\item Joint control and message – NRPGazeboGrpcJointPlugin

This plugin is used to register GazeboJointDataPack DataPack and in this case, only those joints that are explicitly named in the plugin will be registered and made available to control under NRP. The joint’s name must be unique and once again in the plugin declared. In contrast to the other plugins described above or below, when using NRPGazeboGrpcJointPlugin DataPacks can be used to set a target state for the referenced joint, the plugin is integrated with the PID controller and can for each of the joint-specific set a better control performance. 

This plugin must be declared under the corresponding <model> label and have the parallel level in contrast to the <joint> label, and there are four joints that would be chosen to control: rear left and right wheel joint, front left and right steering joint, and according to small tests of the physical model of experiment-vehicle in Gazebo are the parameters of PID controller listed in below block:

\begin{lstlisting}[language=xml]
<model name='smart_car'>  
   ...  
   <joint name="rear_left_wheel_joint">...</joint>  
   <joint name="rear_right_wheel_jointt">...</joint>  
   <joint name="front_left_steering_joint">...</joint>  
   <joint name="front_right_steering_joint">...</joint>  
   ...  
   <plugin name='smart_car_joint_plugin' filename='NRPGazeboGrpcJointControllerPlugin.so'>  
      <rear_left_wheel_joint P='10' I='0' D='0' Type='velocity' Target='0' IMax='0' IMin='0'/>  
      <rear_right_wheel_joint P='10' I='0' D='0' Type='velocity' Target='0' IMax='0' IMin='0'/>  
      <front_left_steering_joint P='40000.0' I='200.0' D='1.0' Type='position' Target='0' IMax='0' IMin='0'/>  
      <front_right_steering_joint P='40000.0' I='200.0' D='1.0' Type='position' Target='0' IMax='0' IMin='0'/>  
   </plugin>  
    ...  
</model>  
\end{lstlisting}

\textbf{Attention:} There are two target types that can be influenced and supported in Gazebo: Position and Velocity. And for the rear left and right wheels of the vehicle are recommended for setting type with “Velocity” and for the front left and right steering are recommended setting type with “Position”. Because the actual control of the rear wheels is better with velocity and front steering uses angle to describe the turning control.

\item Gazebo link information – NRPGazeboGrpcLinkPlugin

This plugin is used to register GazebolinkDataPack DataPacks for each link of the experiment vehicle. Similar to the sensor plugin, this plugin must be declared under <model> label and has the parallel level of <link> label, and only be declared once:

\begin{lstlisting}[language=xml]
<model name='smart_car'>  
   ...  
   <plugin name='smart_car_link_plugin' filename='NRPGazeboGrpcLinkControllerPlugin.so'/>  
   ...  
   <link name='base_link'>...</link>  
   <link name='eye_vision_camera'>...</link>  
   <link name='front_left_steering_link'>...</link>  
   <link name='front_left_wheel_link'>...</link>  
   <link name='front_right_steering_link'>...</link>  
   <link name='front_right_wheel_link'>...</link>  
   <link name='rear_left_wheel_link'>...</link>  
   <link name='rear_right_wheel_link'>...</link>  
    ...  
</model>
\end{lstlisting}

\end{itemize}

\subsubsection{State Transceiver Function ``state\_tf.py”}

State Transceiver Function acquires the location information from the Gazebo engine and transmits it to Vehicle Control Engine to compute the next control commands. The receiving of location coordinates of the vehicle is based on the DataPack from Gazebo, and this DataPack is already encapsulated in NRP, it only needs to in the Decoder indicate which link information should be loaded in DataPack.

\begin{lstlisting}[language=python]
@EngineDataPack(keyword='state_gazebo', id=DataPackIdentifier('smart_car_link_plugin::base_link', 'gazebo')) 
@TransceiverFunction("car_ctl_engine")  
def car_control(state_gazebo):
\end{lstlisting}

The location coordinates in the experiment would be the coordinate of base-chassis “base\_link” chosen and use C++ inheritance declaration with the name of the plugin that is declared in the SDF file. And the received DataPack with the user-defined keyword “state\_gazebo” would be sent in Transceiver Function “car\_control()”.

\textbf{Attention:} Guarantee to get link-information from Gazebo it is recommended new declaring on the top of the script with the below sentence:

\begin{lstlisting}[language=python]
from nrp_core.data.nrp_protobuf import GazeboLinkDataPack
\end{lstlisting}
that could let NRP accurately communicate with Gazebo.

The link-information DataPack in NRP would be called GazeboLinkDataPack. And its Attributes are listed in next Table~\ref{t1}. In Project are “position” and “rotation” information chosen and set to the “car\_ctl\_engine” engine defining Json DataPack, in the last “return” back to “car\_ctl\_engine”. Use the “JsonDataPack” function to get in other engine-defined DataPack and itself form and assign the corresponding parameter with received information from Gazebo.

\begin{lstlisting}[language=python]
car_state = JsonDataPack("state_location", "car_ctl_engine")  
  
car_state.data['location_x'] = state_gazebo.data.position[0]  
car_state.data['location_y'] = state_gazebo.data.position[1]  
car_state.data['qtn_x'] = state_gazebo.data.rotation[0]  
car_state.data['qtn_y'] = state_gazebo.data.rotation[1]  
car_state.data['qtn_z'] = state_gazebo.data.rotation[2]  
car_state.data['qtn_w'] = state_gazebo.data.rotation[3]
\end{lstlisting}

\textbf{Tip:} The z-direction coordinate is not necessary. So only x- and y-direction coordinates are included in DataPack to make the size of JSON DataPack smaller and let the transmission more efficient.

\begin{table}[ht]
    \centering
    \begin{tabular}{c|c|c|c}
        \hline
        \hline
         Attribute & Description & Python Type & C Type\\
         \hline
         pos	& Link Position	& numpy.array(3, numpy.float32) & std::array<float,3>\\
         rot	& Link Rotation as quaternion &	numpy.array(4, numpy.float32) &	std::array<float,4> \\
         lin\_vel &	Link Linear Velocity &	numpy.array(3, numpy.float32) &	std::array<float,3>\\
         ang\_vel	& Link Angular Velocity	& numpy.array(3, numpy.float32)	& std::array<float,3>\\
         \hline
         \hline
    \end{tabular}
    \caption{GazeboLinkDataPack Attributes. \textbf{Tip:} the rotation information from Gazebo is quaternion and its four parameters sort sequence is “x, y, z, w”.}
    \label{t1}
\end{table}

\subsubsection{Vehicle Control Engine “car\_ctl\_engine.py”}

The Vehicle Control Engine would be written according to the form of Python Json Engine. The construction of a Python Json Engine is similar to the definition of a python class file that includes the attributes such as parameters or initialization and its functions. And a class file should declare that this Python Json Engine inherits the class “EngineScript” to let NRP recognize this file as a Python Json Engine to compute and execute. So a Python Json Engine can mostly be divided into three main blocks with def functions: def initialize(self), def runLoop(self, timestep\_ns), and def shutdown(self). 

\begin{itemize}
\item In \textbf{initialize block} is the initial parameters and functions defined for the next simulation. And in this block, should the corresponding DataPacks that belong to the specific Engine at the same time be defined with “self.\_registerDataPack()” and “self.\_setDataPack()” functions:

\begin{lstlisting}[language=python]
self._registerDataPack("actors")  
self._setDataPack("actors", {"angular_L": 0, "angular_R": 0, "linear_L": 0, "linear_R": 0})  
self._registerDataPack("state_location")  
self._setDataPack("state_location", { "location_x": 0, "location_y": 0, "qtn_x": 0, "qtn_y": 0,"qtn_z": 0,"qtn_w": 0})
\end{lstlisting}

    \begin{itemize}
    \item \_registerDataPack():  - given the user-defined DataPack in the corresponding Engine.
    \item \_setDataPack(): 	- given the corresponding name of DataPack and set parameters, form, and value of the DataPack.
    \end{itemize}

The generated actors-control-commands and location-coordinate of the vehicle in this project would be as properties of the DataPack belonging to the “car\_ctl\_engine” Engine.

\item \textbf{runLoop} block is the main block that would always be looped during the simulation progress, which means the computation that relies on time and always need to update would be written in this block. In the “car\_ctl\_engine” Engine should always get the information from Gazebo Engine with the function “self.\_getDataPack()”:

\begin{lstlisting}[language=python]
state = self._getDataPack("state_location") 
\end{lstlisting}

    \begin{itemize}
    \item \_getDataPack():			- given the user\_defined name of the DataPack
    
    \textbf{Attention:} the name must be same as the name in the Transceiver function that user-chosen DataPack which is sent back to Engine.
    \end{itemize}

After the computation of the corresponding command to control the vehicle is the function “\_setDataPack()” once again called to set the commands information in corresponding “actors” DataPack and waiting for other Transceiver Function to call this DataPack:

\begin{lstlisting}[language=python]
self._setDataPack("actors", {"angular_L": steerL_angle, "angular_R": steerR_angle, "linear_L": rearL_omiga, "linear_R": rearR_omiga}) 
\end{lstlisting}

\item \textbf{shutdown} block is only called when the simulation is shutting down or the Engine arises errors and would run under progress.

\end{itemize}

\subsubsection{Package of Euler-angle-quaternion Transform and Trajectory}

\begin{itemize}
\item Euler-angle and quaternion transform

The received information of rotation from Gazebo is quaternion. That should be converted into Euler-angle to conveniently compute the desired steering angle value according to the beforehand setting trajectory. And this package is called “euler\_from\_quaternion.py” and should be in the 
 “car\_ctl\_engine” Engine imported.

\item Trajectory and Computation of target relative steering angle

The beforehand setting trajectory consists of many equal proportional divided points-coordinate. And through the comparison of the present location coordinate and the target coordinate, the package would get the desired distance and steering angle to adjust whether the vehicle arrives at the target. If the vehicle arrives in the radius 0.8m of the target location points will be decided that the vehicle will reach the present destination, and the index will jump to the next destination location coordinate until the final destination. This package is called “relateAngle\_computation.py”.
\end{itemize}

\subsubsection{Actors “Motor” Setting Transceiver Function “motor\_set\_tf.py”}

This Transceiver Function is the communication medium similar to the state-Transceiver Function. The direction of data is now from the “car\_ctl\_engine” Engine to the Gazebo engine. The acquired data from the “car\_ctl\_engine” Engine is the DataPack “actors” with the keyword “actors”:

\begin{lstlisting}[language=python]
@EngineDataPack(keyword='actors', id=DataPackIdentifier('actors', 'car_ctl_engine'))  
@TransceiverFunction("gazebo")  
def car_control(actors):  
\end{lstlisting}

And the DataPack from the Gazebo joint must be validated in this Transceiver Function with the “GazeboJointDataPack()” function. This function is specifically provided by Gazebo to control the joint, the given parameters are the corresponding joint name (declared with NRPGazeboGrpcJointPlugin plugin name in the SDF file) and target Gazebo engine (gazebo) (Attention: each joint should be registered as a new joint DataPack):

\begin{lstlisting}[language=python]
rear_left_wheel_joint = GazeboJointDataPack("smart_car_joint_plugin::rear_left_wheel_joint", "gazebo")  
rear_right_wheel_joint = GazeboJointDataPack("smart_car_joint_plugin::rear_right_wheel_joint", "gazebo")  
front_left_steering_joint = GazeboJointDataPack("smart_car_joint_plugin::front_left_steering_joint", "gazebo")  
front_right_steering_joint = GazeboJointDataPack("smart_car_joint_plugin::front_right_steering_joint", "gazebo") 
\end{lstlisting}

The joint control DataPack is GazeboJointDataPack and its attributes are listed in Table~\ref{t2}:

\begin{table}[ht]
    \centering
    \begin{tabular}{c|c|c|c}
        \hline
        \hline
         Attribute & Description & Python Type & C Type\\
         \hline
         position	& Joint angle position (in rad)	& float & float\\
         velocity	& Joint angle velocity (in rad/s) &	float &	float \\
         effort &	Joint angle effort (in N) &	float &	float\\
         \hline
         \hline
    \end{tabular}
    \caption{GazeboJointDataPack Attributes.}
    \label{t2}
\end{table}

\textbf{Attention:} Guarantee to send Joint-information to Gazebo it is recommended new declaring on the top of the script with the below sentence:

\begin{lstlisting}[language=python]
from nrp_core.data.nrp_protobuf import GazeboJointDataPack
\end{lstlisting}

\subsubsection{Camera Frame-Image Transceiver Function “camera\_tf.py”}

Camera frame-image Transceiver Function acquires the single frame image gathered by Gazebo internally installed camera plugin and sends this frame image to YOLO v5 Engine “yolo\_detector”. The receiving of the image of the camera is based on the camera DataPack from Gazebo called “GazeboCameraDataPack”. To get the data, should the Decorator declare the corresponding sensor name with Validation through C++ and indicate the “gazebo” engine and assign a new keyword for the next Transceiver Function:

\begin{lstlisting}[language=python]
@EngineDataPack(keyword='camera', id=DataPackIdentifier('smart_camera::camera', 'gazebo'))  
@TransceiverFunction("yolo_detector")  
def detect_img(camera):
\end{lstlisting}

\textbf{Attention:} Guarantee to acquire camera information from Gazebo it is recommended new declaring on the top of the script with the below sentence that confirms import GazeboCameraDataPack:

\begin{lstlisting}[language=python]
from nrp_core.data.nrp_protobuf import GazeboCameraDataPack
\end{lstlisting}

And received image Json-information is four parameters: height, width, depth, and image data. The Attributes of the GazeboCameraDataPack are listed in Table~\ref{t3}:

\begin{table}[ht]
    \centering
    \begin{tabular}{c|c|c|c}
        \hline
        \hline
         Attribute & Description & Python Type & C Type\\
         \hline
         image\_height & Camera Image height	& uint32 & uint32\\
         image\_width & Camera Image width &	uint32 &	uint32 \\
         image\_depth &	\makecell[c]{Camera Image depth. \\ Number of bytes per pixel} &	uint8 &	uint32\\
         image\_data & \makecell[c]{Camera Image data. \\ 1-D array of pixel data} & \makecell[c]{numpy.array(image\_height \\ * image\_width * image\_depth, \\ numpy.uint8)} & std::vector<unsigned char>\\
         \hline
         \hline
    \end{tabular}
    \caption{GazeboCameraDataPack Attributes.}
    \label{t3}
\end{table}

The received image data from the gazebo is a 1-D array of pixels with unsigned-int-8 form in a sequence of 3 channels. So this Transceiver Function should be pre-processed with NumPy “frombuffer()” function that transforms the 1-D array in NumPy form:

\begin{lstlisting}[language=python]
imgData = np.frombuffer(trans_imgData_bytes, np.uint8)
\end{lstlisting}

And in the end, validate the Json-DataPack from YOLO v5 Engine and set all information in DataPack, and return to YOLO v5 Engine:

\begin{lstlisting}[language=python]
processed_image = JsonDataPack("camera_img", "yolo_detector")  
  
processed_image.data['c_imageHeight'] = trans_imgHeight  
processed_image.data['c_imageWidth'] = trans_imgWidth  
processed_image.data['current_image_frame'] = imgData 
\end{lstlisting}

\subsubsection{YOLO v5 Engine for Detection of the Objects “yolo\_detector\_engine.py”}

YOLO v5 Engine acquires the camera frame image from Gazebo during the camera Transceiver Function and detects objects in the current frame image. In the end, through the OpenCV package, the result is shown in another window. And the Yolo v5 Engine is also based on the Python Json Engine model and is similar to the vehicle control Engine in section 3.4.2. The whole structure is divided into three main blocks with another step to import Yolo v5 package.

\begin{itemize}

\item Initialization of Engine with establishing “camera\_img” DataPack and validation Yolo v5 object with specific pre-preparation by “detectorWarmUp()”: 
\begin{lstlisting}[language=python]
self._registerDataPack("camera_img")  
self._setDataPack("camera_img", {"c_imageHeight": 0, "c_imageWidth": 0, "current_image_frame": [240 , 320 , 3]})  
self.image_np = 0  
  
self.detector = Yolov5.Yolov5Detector()  
stride, names, pt, jit, onnx, engine, imgsz, device = self.detector.detectorInit()  
self.detector.detectorWarmUp()  
\end{lstlisting}

\item In the main loop function first step is to acquire the camera image with the “\_getDataPack()” function. And the extracted image data from Json DataPack during the camera Transceiver Function became already again in 1-D “list” data form. There is a necessary step to reform the structure of the image data to fit the form for OpenCV. The first is to convert the 1-D array into NumPy ndarray form and, according to acquired height and width information, reshape this np-array. And image form for OpenCV is the default in “BGR” form, and the image from Gazebo is “RGB”. There is also an extra step to convert the “RGB” shaped NumPy ndarray~\cite{opencv}. In the last, it sends the original NumPy array-shaped image and OpenCV-shaped image together into detect-function and finally returns an OpenCV-shaped image with an object-bonding box, and this OpenCV-shaped ndarray can directly use the function of OpenCV showed in the window:

\begin{lstlisting}[language=python]
# Image conversion   
img_frame = np.array(img_list, dtype=np.uint8)  
cv_image = img_frame.reshape((img_height, img_width, 3))  
cv_image = cv_image[:, :, ::-1] - np.zeros_like(cv_image)  
np_image = cv_image.transpose(2,0,1)  
  
# Image detection by Yolo v5  
cv_ImgRet,detect,_ = self.detector.detectImage(np_image, cv_image, needProcess=True)  
  
# Show of Detected image through OpenCV  
cv2.imshow('detected image', cv_ImgRet)  
cv2.waitKey(1)  
\end{lstlisting}

\end{itemize}

%% file: sections/4_result.tex
\section{Simulation Result and Analysis of Performance}

\begin{figure}[ht]
    \centering
    \subfloat[][]{
    \includegraphics[height=5.9cm]{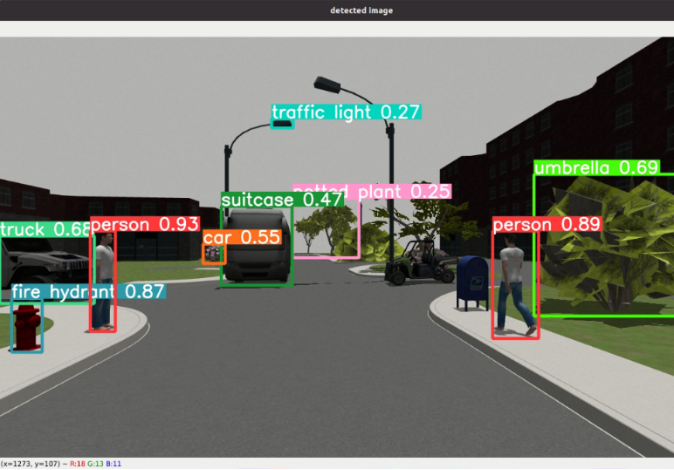}
    }
    \subfloat[][]{
    \includegraphics[height=5.9cm]{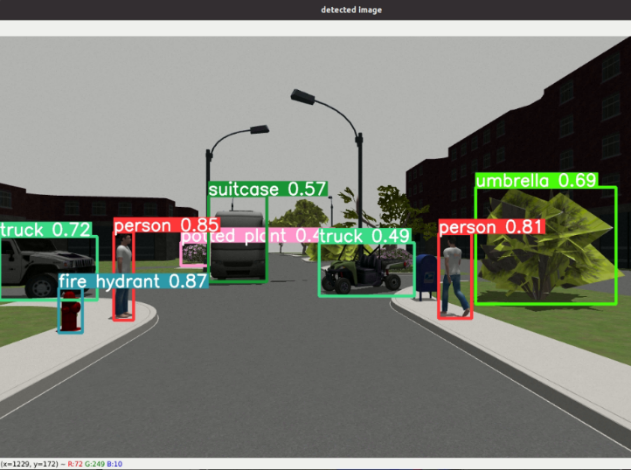}
    }
    
    \caption{Object-detection by Yolo v5 on NRP platform (right: another frame)}
    \label{detection}
\end{figure}

The final goal of the Autonomous driving Benchmark Platform is to build a real-world simulation platform that can train, do research, test or validate different AI algorithms integrated into vehicles, and next, according to the performance to give benchmark and evaluation to adjust algorithms, in the end to real installed these algorithms on the real vehicle. This project “Autonomous Driving Simulator and Benchmark on Neurorobotics Platform” is a basic and tentative concept and foundation to research the possibility of the simulator with multi-agents on the NRP-Core platform. And according to the above construction of a single vehicle agent, the autonomous driving simulation experiment has been finished. This section will discuss the results and suggestions based on the performance of the simulation on the NRP-Core Platform and the Gazebo simulator.

\subsection{Simulation Result of Object-detection and Autonomous Driving}

\subsubsection{Object Detection through YOLOv5 on NRP}

The object detection is based on the visual camera from the Gazebo simulator through the Yolo v5 algorithm. NRP-Core is the behind transmit medium between the Gazebo and Yolo v5 detector. The simulation result is shown in Fig.~\ref{detection}.

On the point of objects-detection, the result reaches the standard and performances well, most of the objects in the camera frame image has been detected, but in some different frame, the detected objects are not stable and come to “undetected.” And in the other hand, although most objects are correctly detected with a high confidence coefficient, e.g., the person is between 80\% ~ 93\%, at the same time, there are few detected errors, such as when the flowering shrubs are detected as a car or a potted plant, the bush plant is detected as an umbrella and the but in front of the vehicle is detected as a suitcase. And last, even though the Yolo works well on the NRP platform, the performance is actually not smooth, and in the Gazebo simulator, the running frame rate is very low, perhaps only around 10-13 frames per second, in a more complex situation, the frame rate came to only 5 frames per second. That makes the simulation in Gazebo very slow and felled the sense of stumble. And when the size and resolution ratio of the camera became bigger, that made the stumble situation worse.

\subsubsection{Autonomous Driving along pre-defined Trajectory}

\begin{figure}[ht]
    \centering
    \subfloat[][]{
    \includegraphics[height=4.5cm]{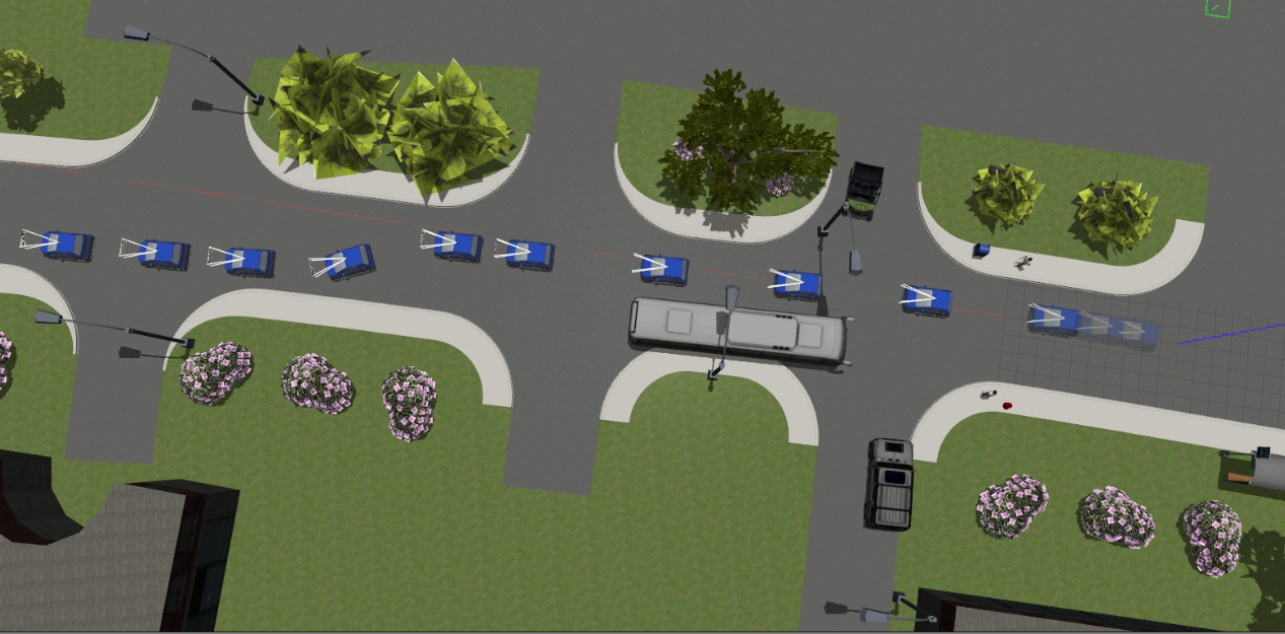}
    }
    \subfloat[][]{
    \includegraphics[height=4.5cm]{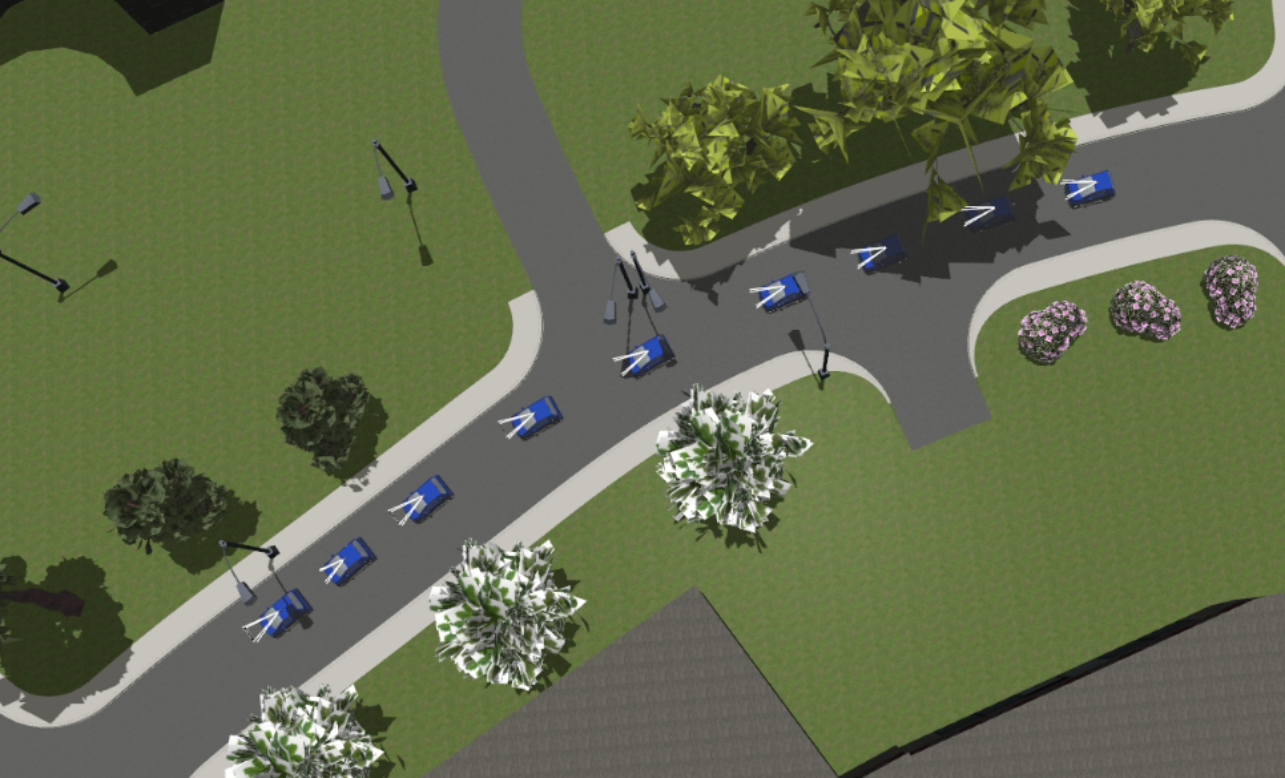}
    }
    
    \caption{Simulation trajectory of autonomous driving}
    \label{trajectory }
\end{figure}

Autonomous driving along a pre-defined trajectory works well, the performance of simulation also runs smoothly and the FPS (frame pro second) holds between 20-40 fps. This FPS ratio is also in the tolerance of real-world simulation. The part trajectory of the experiment vehicle is shown in Fig.~\ref{trajectory }, and the vehicle could run around Parkring and finish one circle. As the first image of the experiment, the vehicle would, according to the detection result, make the corresponding decision to control the vehicle to accelerate or to brake down and turn to evade other obstacles. But for this project, there is no appropriate autonomous driving algorithm to support presently, so here only use a pre-defined trajectory consisting of plenty of point coordinates. The speed of the vehicle is also fixed, and using PID controller to achieve simulated autonomous driving.

And on the other hand, all the 3-D models are equal in proportion to the real size of objects. After many tests of different sizes of the world maps, the size of Parkring is almost the limit of the Gazebo, even though the complexity of the map is not high. For a bigger scenario of the map, the FPS is obviously reduced, and finally, the simulation would become stumble and generate a sense of separation.

\subsubsection{Multi-Engines united Simulation}

The final experiment is to start the Yolo v5 Engine and the autonomous driving control Engine. The above experiments are loaded with only one Engine, and they actually reacted well and had a relatively good performance. And the goal of this project is also to research the possibility of multi-agent simulation. 

The result of multi-Engines simulation actually works in that the Yolo v5 Engine can detect the image and show it in a window and at the same time, the vehicle can move along the trajectory automatically drive. But the simulation performance is not good, and the FPS can only hold between 9 -11 fps. The driving vehicle in Gazebo moves very slowly and not smoothly, and the simulation time has an enormous error compared to the real-time situation.

\subsection{Analysis of Simulation Performance and Discussion}

\subsubsection{YOLOv5 Detection ratio and Accuracy}

Most of the objects near the vehicle in the field of view of the camera have been detected and have high confidence, but there are also some errors appearing during the detection that some objects in as wrong objects are detected, some far objects are detected bus some obvious close objects are not detected. The reason can conclude in two aspects:

1. The employment of the integrated Yolo v5 algorithm is the original version that is not aimed at the specific purpose of this autonomous driving project and has not been trained according to the specific usage. Its network parameters and arts of objects are original and did not use the specific self-own data set, which makes the result actually have a big error between the detected result and expected performance. So that makes the result described in section 4.1.1 that appears some detection error.

2. The accuracy and reality of 3-D models and environment. The object detection algorithm is actually deeply dependent on the quality of the sent image. Here the quality is not about the resolution size but refers to the “reality” of the objects in the image. The original Yolo v5 algorithm was trained based on real-world images, but the camera images from Gazebo actually have enormous distances from real-world images. But the 3-D models and the environment in Gazebo Simulator are relatively very rough, and like cartoon style, they have a giant distance to the real-world objects on the side of the light, material texture of surface and reflection, the accuracy of objects. For example, in Gazebo, the bus has terrible texture and reflection that lets the bus be seen as a black box and not easy to recognize, and Yolo Engine actually detected as a suitcase. And the Environment in Gazebo is also not well exquisitely built. For example, the shrub and bushes on the roadside have a rough appearance with coarse triangles and obvious polygon shapes. That would make huge mistakes and influence the accuracy of desired algorithms.

\begin{figure}[ht]
    \centering
    \subfloat[][]{
    \includegraphics[height=6.5cm]{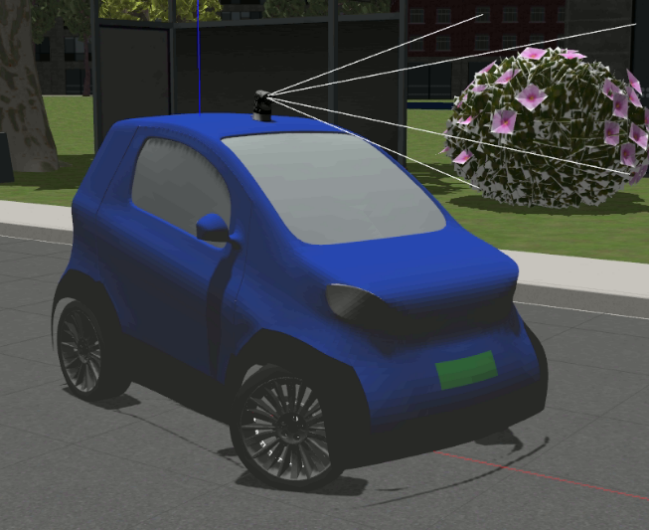}
    }
    \subfloat[][]{
    \includegraphics[height=6.5cm]{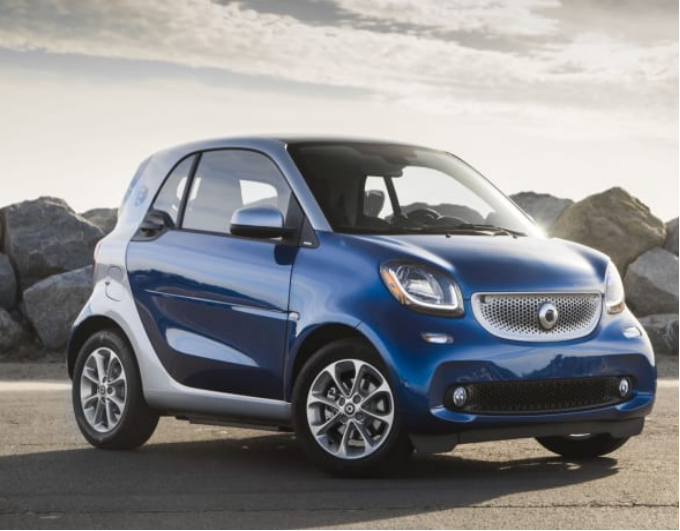}
    }
    
    \caption{Distance between real-world and visual camera image}
    \label{visual}
\end{figure}

3. The property of the Gazebo simulator. The Gazebo simulator is perhaps suitable for small scene simulations like in a room, a tank station, or in a factory. Comparing to other simulators on the market like Unity or Unreal, the advantage of Gazebo is quickly start-up to the reproduction of a situation and environment. But the upper limit of Gazebo and its rendering quality is actually not very close to the real world and can let people at the first time recognize this is a virtual simulation, which also has a huge influence on training object-detection algorithms. And the construction of the virtual world in Gazebo is very difficult and has to use other supported applications like Blender~\cite{blender} to help the construction. Even in Blender, the world has a very high reality, but after the transfer to Gazebo, the rendering quality becomes terrible and awful.

In fact, although detection has some mistakes and errors, the total result and performance are in line with the forecast that the Yolo v5 algorithm has excellent performance.

\subsubsection{Multi-Engines Situation and Non-smooth Simulation Phenomenon}

The simulation of single loaded Yolo Engine and the multi-engine meanwhile operation appear terrible performance by the movement of the vehicle and inferior progress FPS of the whole simulation. But simulation for single loaded vehicle control engine is actually working well and has smooth performance. After the comparison experiment, the main reason for the terrible performance is because of the backstage transmission mechanism of information between Python Json Engine on the NRP Platform. In the simulation of a single loaded vehicle control Engine, the transmission from Gazebo is based on Protobuf-gRPC protocol, and transmission back to Gazebo is JSON protocol, but the size of transmitted information is actually very small because the transmitted data consists of only the control commands like “line-velocity” and “angular-velocity” that don’t take much transmission capacity and for JSON Protocol is actually has a negligible error to Protobuf Protocol. And the image transmission from Gazebo to Transceiver Function is also based on the Protobuf-gRPC method. But the transmission of an image from the Transceiver Function to Yolo Engine through JSON Protocol is very slow because the information of an image is hundreds of commands, and the according to the simulation loop in NRP, would make a block during the process of simulation and let the system “be forced” wait for the finish of transmission of the image. The transfer efficiency of JSON Protocol is actually compared to real-time slowness and tardiness, which takes the choke point to the transmission and, according to the test, only reduces the resolution rate of the camera to fit the simulation speed requirements.

\subsection{Improvement Advice and Prospect}

The autonomous driving simulator and application on NRP-Core achieve the first goal of building a concept and foundation for multi-agents, and at the same time, this model is still imperfect and has many disadvantages that would be improved. On the NRP-Core platform is also the possibility for a real-world simulator discussed, and the NRP-Core has large potential to achieve the complete simulation and online cooperation with other platforms. There are also some directions and advice for the improvement of this application presently on NRP for further development.

\subsubsection{Unhindered simulation with other communication protocol}

As mentioned before, the problem that communication with JSON protocol is the simulation at present is not smooth and has terrible simulation performance with Yolo Engine. Actually, the transmission of information through the Protobuf protocol based on the transmission between Gazebo and Transceiver Functions has an exceeding expectation performance than JSON protocol. The development Group of NRP-Core has also been developing and integrating the Protobuf-gRPC~\cite{protocol} communication backstage mechanism on the NRP-Core platform to solve the big data transmission problem. And in order to use Yolo or other object-detection Engines, it is recommended to change the existing communication protocol in the Protobuf-gRPC protocol. And the Protobuf protocol is a free and open-source cross-platform data format used to serialize structured data and developed by google, and details see on the official website~\cite{protocol}.

\subsubsection{Selection of Basic Simulator with better performance}

\begin{figure}[ht]
    \centering
    \subfloat[][Sunny]{
    \includegraphics[height=4.5cm]{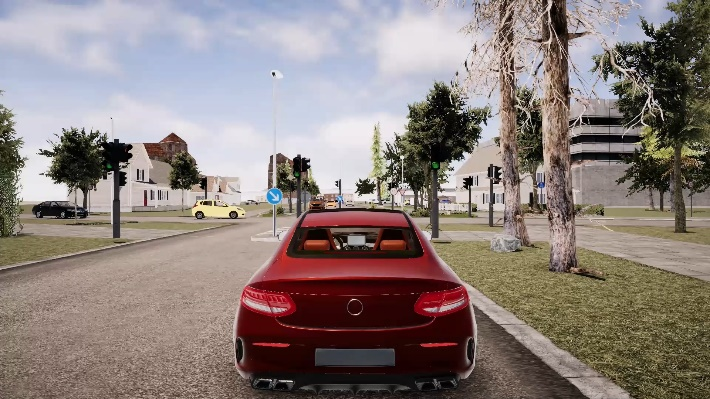}
    }
    \subfloat[][Foggy]{
    \includegraphics[height=4.5cm]{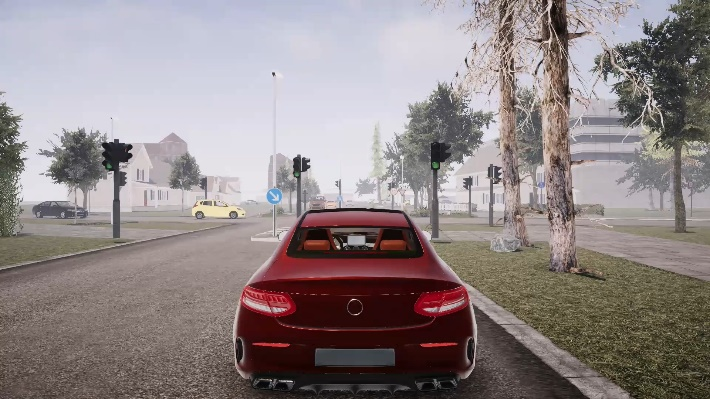}
    }
    \\
    \subfloat[][Raining]{
    \includegraphics[height=4.5cm]{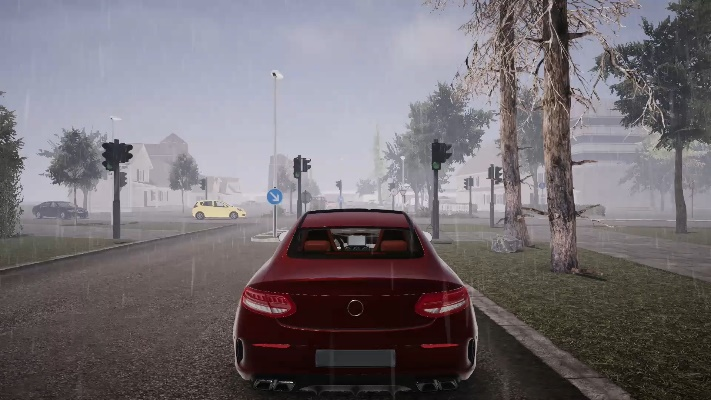}
    }
    \subfloat[][Snowy]{
    \includegraphics[height=4.5cm]{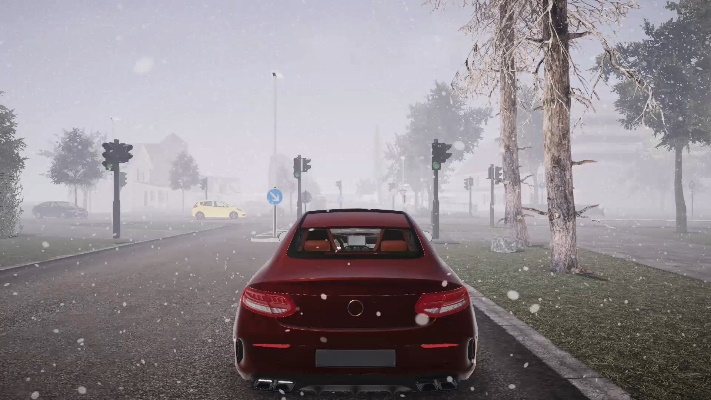}
    }
    \caption{Construction of simulation world in Unity3D with weather application}
    \label{unity}
\end{figure}

Because of the limitation of performance and functions of the Gazebo, there are many applications that can not in Gazebo easy to realize, such as the weather and itself change, and the accuracy and reality of 3-D models also have limitations. The usage of high-accuracy models would make the load became heavier on the Gazebo because of the fall behind the optimization of the Gazebo simulator. In fact, there are many excellent simulators, and they also provide many application development packages that can shorten the development period, such as Unity3D~\cite{unity} or Unreal engine simulator~\cite{ue}. In the team of an autonomous driving simulator and the benchmark there is an application demo on Unity3D simulator and figure Fig.~\ref{unity} shows the difference between Gazebo and Unity3D.

\begin{figure}[ht]
    \centering
    \subfloat[][Detection by YOLOv5 on Gazebo]{
    \includegraphics[height=4.1cm]{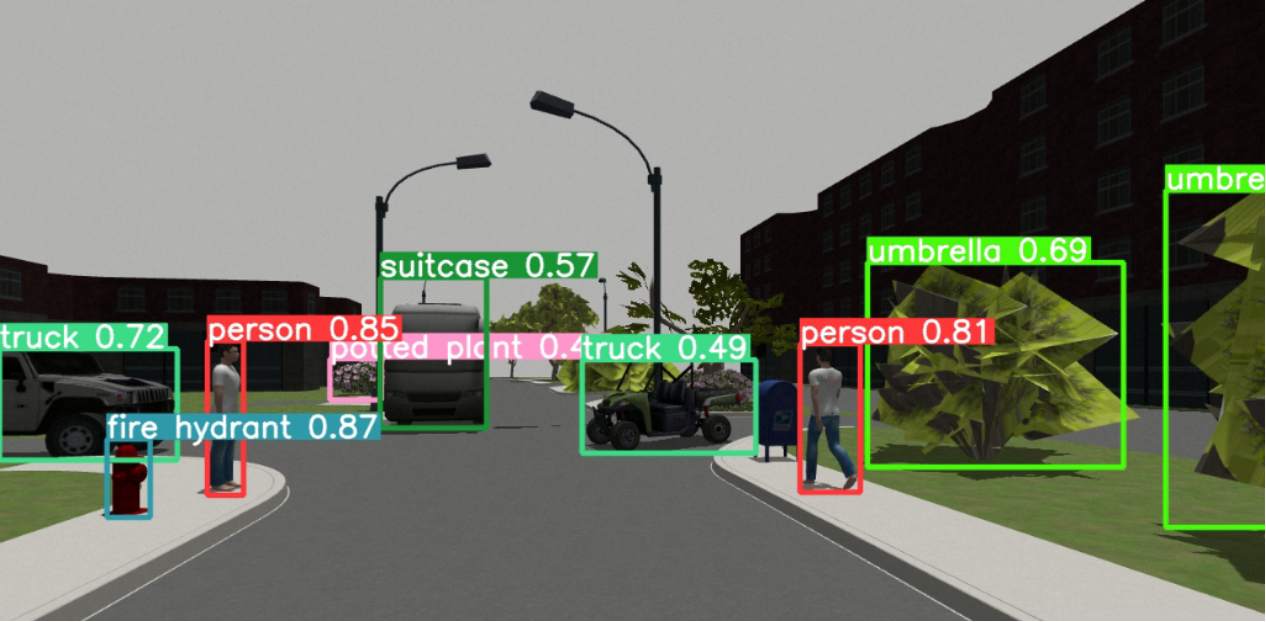}
    }
    \subfloat[][Detection by YOLOv5 on Unity3D]{
    \includegraphics[height=4.1cm]{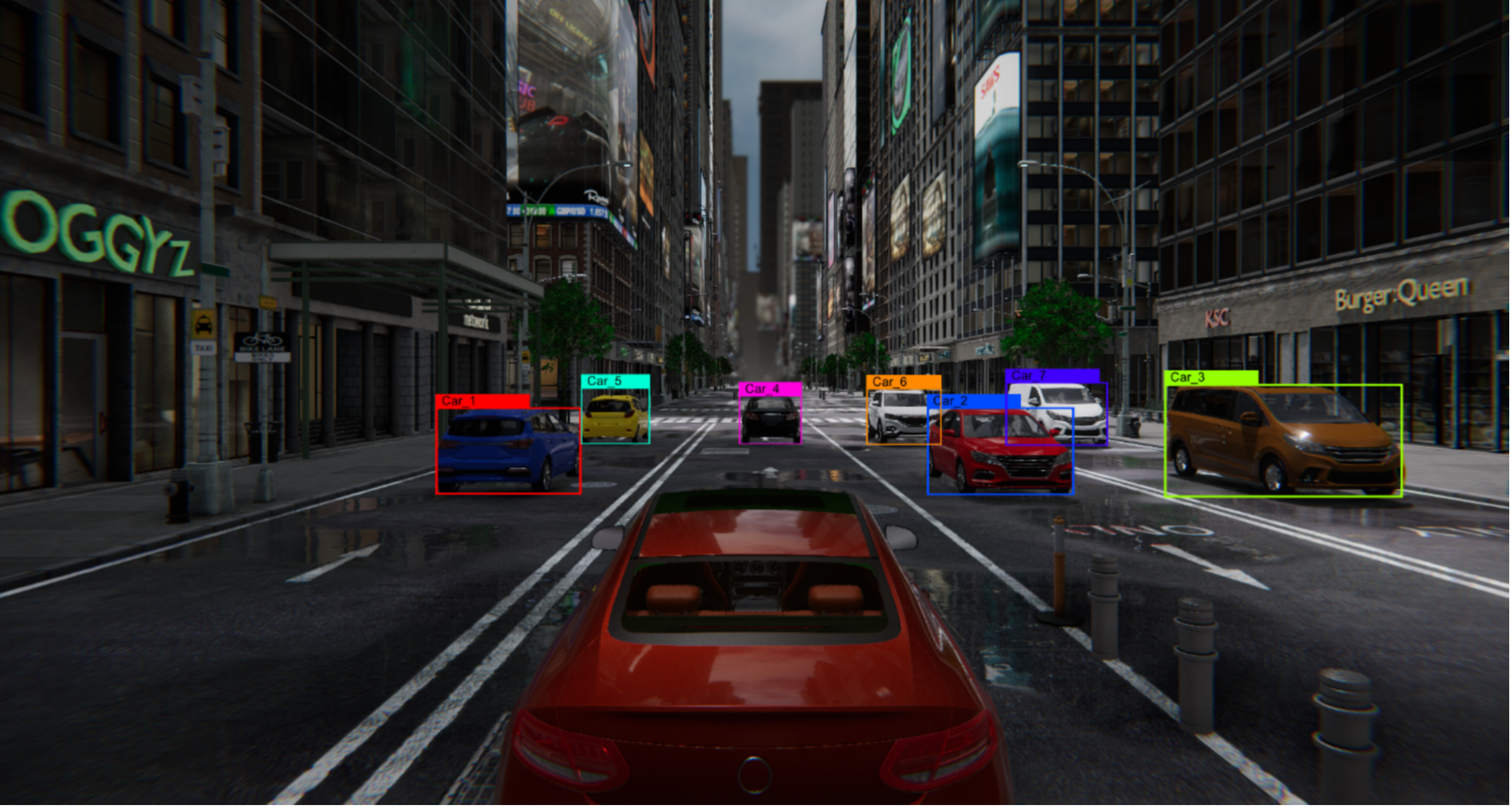}
    }
    \caption{Comparing of the detection result by different platforms}
    \label{gazuni}
\end{figure}

The construction and simulation in Unity3D have much better rendering quality close to the real world than Gazebo, and the simulation FPS can maintain above 30 or even 60 fps. And for the YoloV5 detection result, according to the analysis in section 4.2.1, the result by Unity3D is better than the performance by Gazebo simulator because of more precision 3-D models and better rendering quality of models (Example see Fig.~\ref{gazuni}). The better choice for the development as the basic simulator and world expresser is recommended to develop on Unity3D or other game engines. And actually, NRP-Core will push a new version that integrates the interfaces with Unity3D and could use Protobuf protocol to ensure better performance for a real-world simulation.

\subsubsection{Comparing to other Communication Systems and frameworks}

There are also many communication transmission frameworks and systems that are widely used in academia or business for robot development, especially ROS (Robot Operating System) system already has many applications and development. Actually, ROS has already been widely and mainly used for Robot-development with different algorithms: detection algorithm and computer vision, SLAM (Simultaneous Localization and Mapping) and Motion-control, and so on. ROS has already provided relatively mature and stable methods and schemes to undertake the role of transmitting these necessary data from sensors to the robot’s algorithms and sending the corresponding control command codes to the robot body or actors. But the reason chosen NRP-Core to be the communication system is based on the concepts of Engines and Transceiver Functions.  Compared to ROS or other framework NRP platform has many advantages: This platform is very easy to build multi-agents in simulation and conveniently load in or delete from the configuration of simulation; The management of information is easier to identify than ROS-topics-system; The transmission of information is theoretically more efficient, and modularization and this platform can also let ROS at the same time as parallel transmission method to match and adapt to another systems or simulations. From this viewpoint, the NRP platform generalizes the transmission of data and extends the boundary of the development of the robot, which makes the development more modular and efficient. ROS system can also realize the multi-agents union simulation but is not convenient to manage based on the "topic" system. ROS system is now more suitable for a single agent simulation and the simulation environment. As mentioned before,  the real interacting environment is not easy to realize. But NRP-Core has the potential because that NRP-Core can at the same time run the ROS system and let the agent developed based on the ROS system easily join in the simulation. That is meaningful to develop further on the NRP-Core platform.

%% file: sections/5_conclusion.tex
\section{Conclusion and Epilogue}

This project focuses on the first construction of the basic framework on the Neurorobotics Platform for applying the Autonomous Driving Simulator and Benchmark. Most of the functions including the template of the autonomous driving function and object-detection functions are realized. The part of the benchmark because there are no suitable standards and further development is a huge project regarded as further complete development for the application. 

This project started with researching the basic characters to build a simulation experiment on the NRP-Core Platform. Then the requirements of the construction of the simulation are listed and each necessary component and object of the NRP-Core is given the basic and key understanding and attention. The next step according to the frame of the NRP-Core is the construction of the application of the autonomous driving simulator. Started with establishing the physic model of the vehicle and the corresponding environment in the SDF file, then building the “close loop” - autonomous driving based on PID control along the pre-defined trajectory and finally the “open loop” – objects-detection based on YoloV5 algorithm and successfully achieve the goal to demonstrate the detected current frame image in a window and operated as camera monitor. And at last, the current problems and the points of improvement are listed and discussed in this development document.

And at the same time there are also many problems that should be optimized and solved. At present the simulation application can only regard as research for the probability of the multi-agent simulation. The performance of the scripts has a lot of space to improve, and it is recommended to select a high-performance simulator as the carrier of the real-world simulation. In fact the NRP-Core platform has shown enormous potential for the construction of a simulation world with each object interacting function and the high efficiency to control and manage the whole simulation project. In conclusion the NRP-Core platform has great potential to achieve the multi-agents simulation world.

%% file: sections/6_appendix.tex
\section{Appendix}

\begin{table}[ht]
    \centering
    \begin{tabular}{@{\hspace{0pt}}m{3.5cm}<{\centering}@{\hspace{0pt}}|@{\hspace{0pt}}m{5cm}<{\centering}@{\hspace{0pt}}|@{\hspace{0pt}}m{3.2cm}<{\centering}@{\hspace{0pt}}|@{\hspace{0pt}}m{1.7cm}<{\centering}@{\hspace{0pt}}|@{\hspace{0pt}}m{1cm}<{\centering}@{\hspace{0pt}}|@{\hspace{0pt}}m{2.2cm}<{\centering}@{\hspace{0pt}}}
    \hline
    \hline
    Name &	Description	& Type &	Default	& Array &	Values  \\
    \hline
    SimulationLoop &	Type of simulation loop used in the experiment &	enum &	"FTILoop" && "FTILoop" "EventLoop" \\
    \hline
    SimulationTimeout &	Experiment Timeout (in seconds). It refers to simulation time &	integer &	0 && \\
    \hline
    SimulationTimestep &	Time in seconds the simulation advances in each Simulation Loop. It refers to simulation time. &	number &	0.01 && \\	
    \hline
    ProcessLauncherType	& ProcessLauncher type to be used for launching engine processes &	string &	Basic && \\	
    \hline
    EngineConfigs &	Engines that will be started in the experiment &	EngineBase &&		X & \\
    \hline
    DataPackProcessor &	Framework used to process and rely datapack data between engines. Available options are the TF framework (tf) and Computation Graph (cg) &	enum &	"tf" &&		"tf", "cg" \\
    \hline
    DataPackProcessing-Functions &	Transceiver and Preprocessing functions that will be used in the experiment &	TransceiverFunction &&		X & \\
    \hline
    StatusFunction &	Status Function that can be used to exchange data between NRP Python Client and Engines &	StatusFunction &&& \\		
    \hline
    ComputationalGraph &	List of filenames defining the ComputationalGraph that will be used in the experiment &	string &&		X & \\
    \hline
    EventLoopTimeout &	Event loop timeout (in seconds). 0 means no timeout. If not specified 'SimulationTimeout' is used instead &	integer &	0 && \\
    \hline
    EventLoopTimestep &	Time in seconds the event loop advances in each loop. If not specified 'SimulationTimestep' is used instead &	number &	0.01 && \\	
    \hline
    ExternalProcesses &	Additional processes that will be started in the experiment &	ProcessLauncher &&		X & \\
    \hline
    ConnectROS &	If this parameter is present a ROS node is started by NRPCoreSim &	ROSNode &&& \\	
    \hline
    ConnectMQTT &	If this parameter is present an MQTT client is instantiated and connected &	MQTTClient &&& \\
    \hline
    \hline
    \end{tabular}
    \caption{Simulation configuration}
    \label{a1}
\end{table}

\begin{table}[ht]
    \centering
    \begin{tabular}{@{\hspace{0pt}}m{4cm}<{\centering}@{\hspace{0pt}}|@{\hspace{0pt}}m{5.5cm}<{\centering}@{\hspace{0pt}}|@{\hspace{0pt}}m{1.5cm}<{\centering}@{\hspace{0pt}}|@{\hspace{0pt}}m{2.5cm}<{\centering}@{\hspace{0pt}}|@{\hspace{0pt}}m{1.5cm}<{\centering}@{\hspace{0pt}}|@{\hspace{0pt}}m{1cm}<{\centering}@{\hspace{0pt}}}
    \hline
    \hline
    Name &	Description &	Type &	Default & Required &	Array \\
    \hline
    EngineName &	Name of the engine &	string &&		X & \\
    \hline
    EngineType &	Engine type. Used by EngineLauncherManager to select the correct engine launcher &	string &&	X &	\\
    \hline
    EngineProcCmd &	Engine Process Launch command &	string &&& \\
    \hline
    EngineProcStartParams &	Engine Process Start Parameters	& string &	[ ]		&& X \\
    \hline
    EngineEnvParams &	Engine Process Environment Parameters &	string &	[ ] &&		X \\
    \hline
    EngineLaunchCommand &	LaunchCommand with parameters that will be used to launch the engine process &	object &	{"LaunchType": "BasicFork"} && \\
    \hline
    EngineTimestep &	Engine Timestep in seconds &	number &	0.01 && \\
    \hline
    EngineCommandTimeout &	Engine Timeout (in seconds). It tells how long to wait for the completion of the engine runStep. 0 or negative values are interpreted as no timeout &	number &	0.0	&& \\
    \hline
    \hline
    \end{tabular}
    \caption{Engine Base Parameter}
    \label{a2}
\end{table}